\documentclass[conference]{IEEEtran}
\IEEEoverridecommandlockouts

\usepackage{cite}
\usepackage{amsmath,amssymb,amsfonts}
\usepackage{bm}

\usepackage{algorithmic}
\usepackage{algorithm}

\usepackage{graphicx}
\usepackage{textcomp}
\usepackage{xcolor}

\usepackage[centerlast,small,sc]{caption}

\usepackage{multirow}
\usepackage{subfig}

\begin{document}

\title{Deep Auto-Set: A Deep Auto-Encoder-Set Network for Activity Recognition Using Wearables}

\author{
	\IEEEauthorblockN{Alireza Abedin Varamin}
	\IEEEauthorblockA{
		\textit{The University of Adelaide}\\
		Adelaide SA 5005, Australia \\
		alireza.abedinvaramin@adelaide.edu.au}
	\and
	\IEEEauthorblockN{Ehsan Abbasnejad}
	\IEEEauthorblockA{
		\textit{The University of Adelaide}\\
		Adelaide SA 5005, Australia \\
		ehsan.abbasnejad@adelaide.edu.au}
	\and
	\IEEEauthorblockN{Qinfeng Shi}
	\IEEEauthorblockA{
		\textit{The University of Adelaide}\\
		Adelaide SA 5005, Australia \\
		javen.shi@adelaide.edu.au}
	\and
	\IEEEauthorblockN{~~~~~~~~~~~~~~~~~~~~~Damith Ranasinghe}
	\IEEEauthorblockA{
		\textit{~~~~~~~~~~~~~~~~~~~~~The University of Adelaide}\\
		~~~~~~~~~~~~~~~~~~~~~Adelaide SA 5005, Australia \\
		~~~~~~~~~~~~~~~~~~~~~damith.ranasinghe@adelaide.edu.au}
	\and
	\IEEEauthorblockN{Hamid Rezatofighi}
	\IEEEauthorblockA{
		\textit{The University of Adelaide}\\
		Adelaide SA 5005, Australia \\
		hamid.rezatofighi@adelaide.edu.au}
}

\maketitle

\begin{abstract}
Automatic recognition of human activities from time-series sensor data (referred to as HAR) is a growing area of research in ubiquitous computing. Most recent research in the field adopts supervised deep learning paradigms to automate extraction of intrinsic features from raw signal inputs and addresses HAR as a multi-class classification problem where detecting a single activity class within the duration of a sensory data segment suffices. However, due to the innate diversity of human activities and their corresponding duration, no data segment is guaranteed to contain sensor recordings of a single activity type. In this paper, we express HAR more naturally as a \textit{set prediction problem} where the predictions are \textit{sets} of ongoing activity elements with unfixed and unknown cardinality. For the first time, we address this problem by presenting a novel HAR approach that learns to output activity sets using deep neural networks. Moreover, motivated by the limited availability of annotated HAR datasets as well as the unfortunate immaturity of existing unsupervised systems, we complement our supervised set learning scheme with a prior unsupervised feature learning process that adopts convolutional auto-encoders to exploit unlabeled data. The empirical experiments on two widely adopted HAR datasets demonstrate the substantial improvement of our proposed methodology over the baseline models.
\end{abstract}

\begin{IEEEkeywords}
Activity Recognition, Deep Learning, Time-series Data, Wearable Sensors
\end{IEEEkeywords}

\section{Introduction}

With the proliferation of low-cost sensing technologies as well as the fast advancements in machine learning techniques, automatic human activity recognition (HAR) using wearable sensors has emerged as a key research area in ubiquitous computing~\cite{Bao:2004,Ordonez:2016,Hammerla:2016,yao2018efficient}. In this problem, high-level activity information is acquired through analyzing low-level sensor recordings with the goal of providing proactive assistance to users. Having created new possibilities in diverse application domains ranging from health-care monitoring to entertainment industry, HAR has successfully sparked excitement in both academia and industry. 
Nevertheless, due to the inherent diverse nature of human activities, HAR faces unique methodological challenges such as intra-class variability, inter-class similarity, class imbalance, the Null class problem~\cite{Bulling:2014}, the multi-class window problem~\cite{yao2018efficient}, and intermittent activity recognition problem~\cite{Kim:2010} to name a few. Accordingly, it is of great significance to propose systematic approaches towards accurate recognition of activities that triumph over the challenges. 

While previous studies have explored both shallow and deep architectures for a diverse range of HAR application scenarios, multi-class classification has been their dominant approach for formulating the problem. As such, sensor time segments obtained from striding a fixed-size sliding window over the sensor data-streams are assigned a single activity class, approximated based on the most \cite{Yang:2015} or the last \cite{Ordonez:2016} observed sample annotations. Such a strategy towards ground-truth approximation is clearly associated with loss of activity information and potentially deludes the supervised training process. This becomes even more problematic since the optimal size for the sliding window is not known a priori~\cite{Bulling:2014} and therefore, no segment is guaranteed to contain measurements of a single activity type~\cite{yao2018efficient}. Equally important, existing deep HAR systems demand large amounts of annotated training data for enhanced supervised performance. However, large-scale annotated HAR datasets are limited. Further, collection of labeled sensory data is labor intensive, time-consuming and expensive \cite{Kim:2010}. As opposed to other domains (\emph{e.g.} image recognition) where human visualization of raw data alleviates the labeling process, manual annotation of sensor signals is a tedious task. Unfortunately, activity recognition systems that leverage the cheaply available unlabeled sensory data are rare in the field and, therefore, necessitates the exploration of effective unsupervised alternatives.   





\begin{figure}[htbp]
	\centering
	\includegraphics[width=80mm, keepaspectratio]{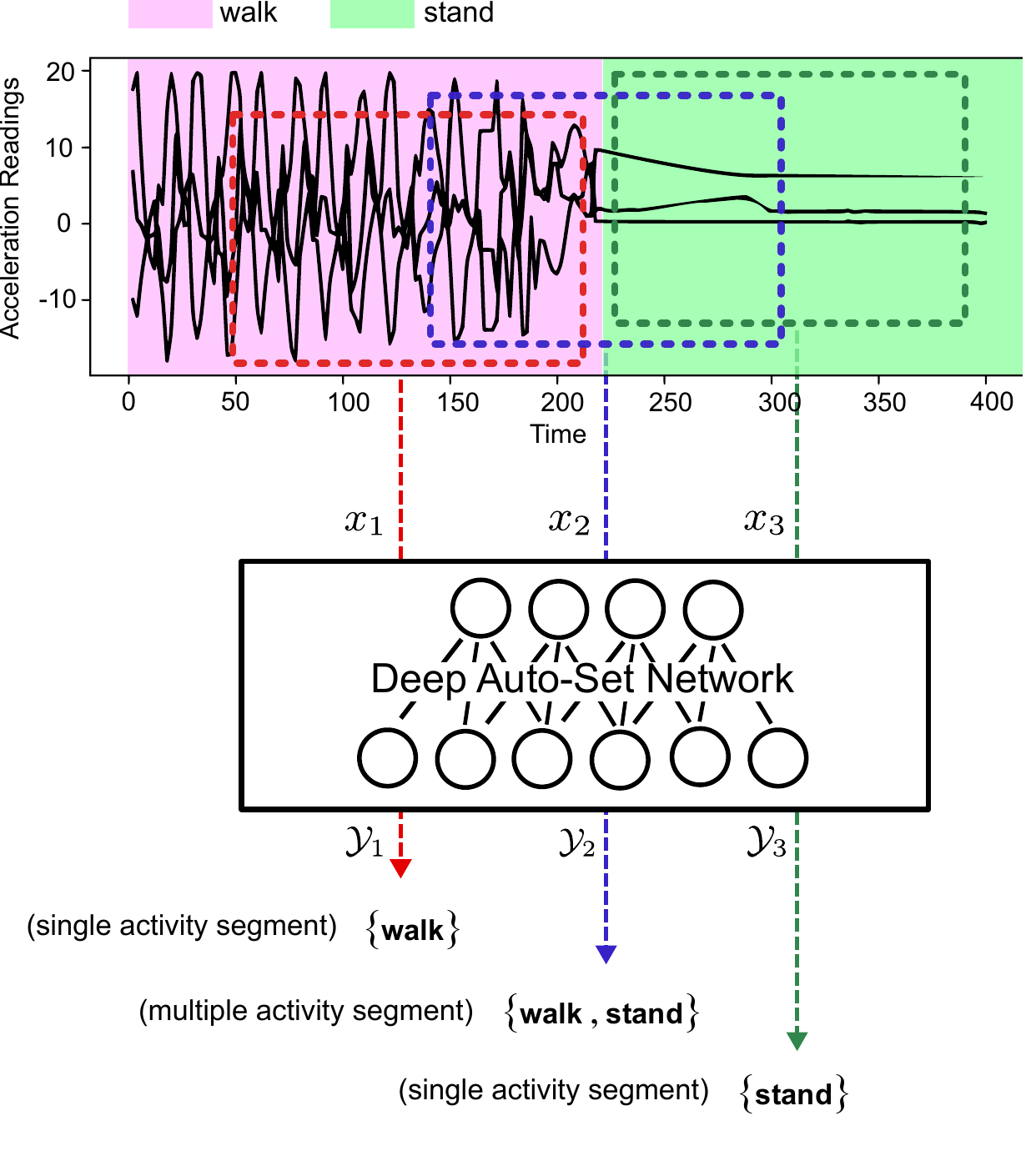}
	\caption{An illustration of our novel \textit{Deep Auto-Set} network to perform precise activity recognition from time-series data. Our network consumes windowed raw sensory excerpts ($\bm{x}$), automatically extracts distinctive features and outputs corresponding \emph{sets of activities} ($\mathcal{Y}$) with various cardinalities.} 
	\label{fig:sliding}
\end{figure}

In this paper, we overcome the innate limitations of multi-class formulated HAR by expressing the problem more naturally as a \textit{set prediction problem}. As such, the goal is to predict the \textit{set} of ongoing activity elements (whose cardinality is unknown and unfixed beforehand) within the duration of a time segment.  For instance, considering a sensory time segment in which the subject of interest is initially walking but then suddenly stops moving, the system is expected to output the set $\{$walk,stand$\}$ to capture the underlying activity transition. Similarly, an output empty set $\{\}$ intuitively expresses a time segment in which the activities of interest did not occur. Inspired by the study in~\cite{Rezatofighi:2017}, for the first time we develop a HAR system that performs activity set learning and inference in a systematic fashion using deep paradigms. In contrast to conventional multi-label approaches, our methodology omits heuristic thresholding methods for selecting activity labels and instead learns to predict cardinality in addition to the activity labels. Further, motivated by the scarcity of annotated HAR datasets, we complement our supervised training scheme with a prior unsupervised feature learning step that exploits unlabeled time-series data. Through experiments on widely adopted public HAR datasets, we demonstrate the significant improvement achieved from proposed deep learning based methodology, the \emph{Deep Auto-Set} network (depicted in Fig. \ref{fig:sliding}), over the baseline models. The main contributions of this paper are summarized as follows: 
\begin{itemize}
	
	\item For the first time, we investigate a novel formulation of a human activity recognition problem from body worn sensor data streams where the predictions for sensory time segments are expressed as \emph{activity sets}. Our novel formulation naturally handles sensory segments with varying number of activities and thus, avoids the potential loss of information from conventional ground-truth approximations.  
	
	\item We present Deep Auto-Set: a unified deep learning paradigm that (a) seamlessly functions on raw multi-modal sensory segments, (b) exploits unlabeled data to uncover effective feature representations, and (c) incorporates set objective to learn mappings from input sensory data to target activity sets. 
	
	\item We demonstrate the effectiveness of our Deep Auto-Set network through empirical experiments on two HAR representative datasets. We further examine the components of our proposed methodology in isolation, to present insights on their contribution to an enhanced recognition performance. 
	
\end{itemize}

\section{Related Work}


\begin{figure*}[h]
	\centering
	\includegraphics[width=\textwidth, keepaspectratio]{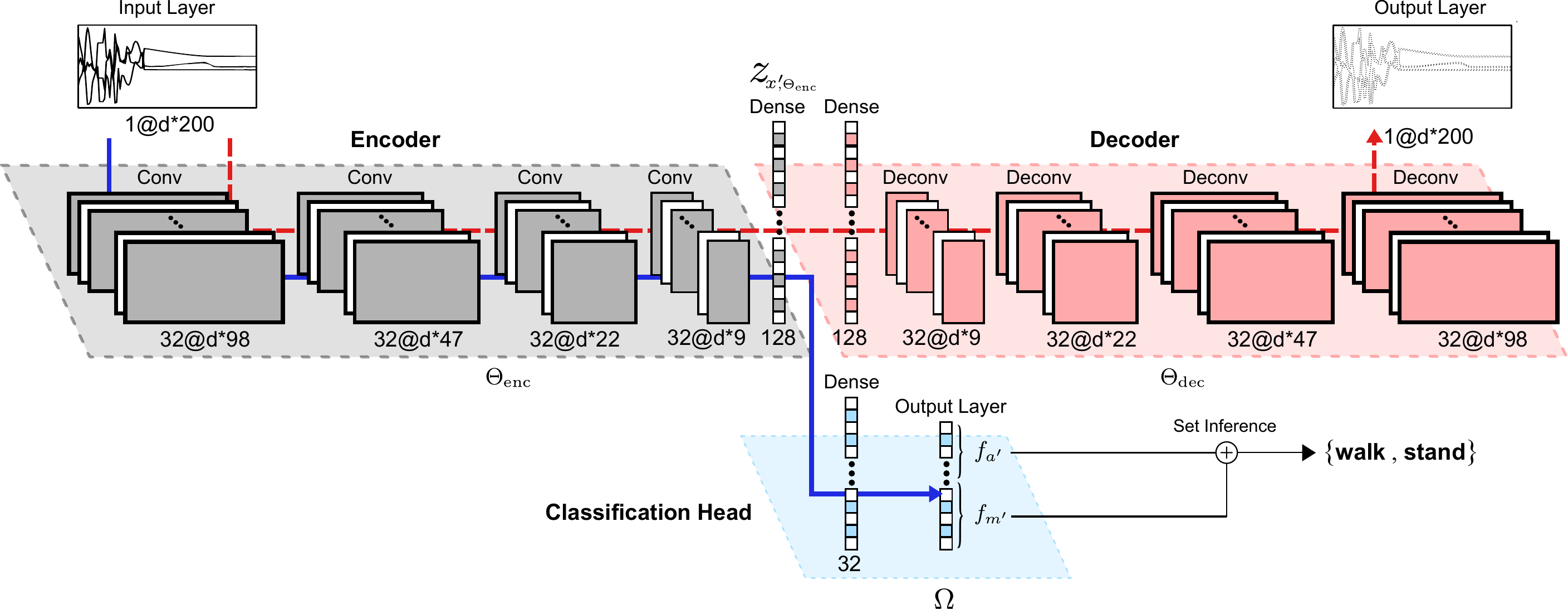}
	\caption{Unified architecture of our \emph{Deep Auto-Set} network. The tags above the feature maps refer to the corresponding layer operations. The numbers before and after "@" respectively correspond to the number of generated feature maps and their dimensions in each layer. In this architecture, all convolution (and deconvolution) layers apply a filter of width 5 (as in \protect\cite{Ordonez:2016}) and stride 2 (for down-sampling) along the temporal dimension of the feature maps. For the unsupervised step, starting from the input layer, layer operations on the dashed arrow are consecutively applied on the generated feature maps of previous layers to output the reconstructed segment; these operations correspond to the convolutional auto-encoder network parameterized by $\Theta_{\text{enc}}$ and $\Theta_{\text{dec}}$. Similarly for the supervised step, operations on the solid arrow correspond to the activity set network parameterized by $\Theta_{\text{enc}}$ and $\Omega$. Once the network parameters are optimized, set inference (as described in Section \ref{sub:sup}) is carried out to generate activity set predictions. }
	\label{fig:Architecture}
\end{figure*}

The well-established activity recognition pipeline for time-series sensory data involves sliding window segmentation, feature extraction, and activity classification \cite{Bulling:2014}. In this regard, adopting hand-crafted features (e.g. statistical features \cite{Ravi:2005}, basis transform features \cite{Huynh:2005}, multi-level features \cite{Zhang:2012}, bio-mechanical features~\cite{wickramasinghe2017sequence}) coupled with employment of shallow classifiers (e.g. support vector machines \cite{Bulling:2012}, decision trees \cite{Bao:2004}, joint boosting \cite{Lara:2012}, graphical models~\cite{ShinmotoTorres:2017CRF}, and multi-layer perceptrons \cite{Randell:2000}) has been extensively explored as the traditional approach to HAR. While this manually tuned procedure has successfully acquired satisfying results for relatively simple recognition tasks, its generalization performance is limited by heavy reliance on domain expert knowledge to design distinctive features. 

Recently, the emerging paradigm of deep learning has presented unparalleled performance in various research areas including computer vision, natural language processing and speech recognition \cite{Lecun:2015}. When applied to sensor-based HAR, deep learning allows for automated end-to-end feature extraction and thus, largely alleviates the need for laborious feature engineering procedures. Motivated by these, we are seeing an increasing  adoption of deep learning paradigms in HAR~\cite{Zeng:2014,Yang:2015,Hammerla:2016,Ordonez:2016,yao2018efficient}. In this regard, convolutional neural networks (CNNs) have appeared as the most popular choice for automatic extraction of effective high-level features. Research in this line includes \cite{Zeng:2014,Yang:2015} where raw sensory data were processed by convolutional layers to extract discriminative features. Going beyond CNNs, Hammerla \emph{et al.} \cite{Hammerla:2016} conducted extensive experiments to investigate suitability of various deep architectures for HAR using wearables and concluded guidelines for hyper-parameter tuning in different application scenarios. Ordóñez and Roggen \cite{Ordonez:2016} developed a recurrent-based neural network (RNN) for wearable sensors and reported state-of-the-art performance on a representative HAR dataset. Except for the dense labelling and prediction approach in~\cite{yao2018efficient}, existing supervised solutions are based on the assumption that all samples within a sliding window segment share the same activity annotation. We argue that such an assumption is counter-intuitive to the diverse nature of human activities with varying durations and hinders accurate analysis of segments with multiple activities. In this paper, we present a novel network that naturally allows segmented sensory data to be associated with a set of activity elements.

Moreover, most existing HAR research solely rely on supervised training for feature extraction. In the absence of sufficiently large annotated datasets, this leads to poor generalization performance. Taking into account the scarcity of annotated HAR datasets and the difficulty of doing so, we exploit unlabeled time-series data to learn useful feature representations by adopting convolutional auto-encoders. In this regard, the most relevant study to ours is \cite{Alsheikh:2016} where layer-wise pre-training of fully connected deep belief networks is adopted and the recognition problem is limited to preprocessed spectrograms of acceleration measurements. In contrast, our proposed unsupervised methodology substitutes the layer-by-layer pre-training with an end-to-end optimization of the reconstruction objective and is also seamlessly applied on raw multi-modal sensor data.

\section{Deep Auto-Set for Human Activity Recognition}\label{method}

Here we elaborate on our novel methodology towards addressing HAR as a set prediction problem, which we refer to as the \textit{Deep Auto-Set}. The working flow of our proposed solution involves an unsupervised feature learning step (described in Section \ref{sub:unsup}) that exploits cheaply accessible unlabeled sensor measurements followed by a supervised fine-tuning step (detailed in Section \ref{sub:sup}) that leverages valuable label information to extract more discriminative features while simultaneously training the network to generate activity sets for the given sensory data. Noting that our methodology is not confined to a specific network architecture, we carry out both supervised and unsupervised tasks by adopting a CNN architecture employed in \cite{Ordonez:2016} as the core of our network and apply modifications to suit our problem settings; this architecture comprises of four convolutional layers followed by two dense layers that apply rectified linear units (ReLUs) for non-linear transformation as well as a softmax logistic regression output layer to yield the classification outcome. 

Specifically for the unsupervised feature learning step, we construct a symmetric convolutional auto-encoder by arranging a chain of deconvolutional operations in the decoder network symmetric to the convolutional layers in the encoder network. This choice is grounded over the success of auto-encoders in improving generalization performance through unsupervised feature learning \cite{Erhan:2010}. 

In addition, for the supervised activity set learning step, the encoder network is augmented with a multi-label classification head and the output layer is adjusted to suit the set formulation. The overall architecture of our \textit{Deep Auto-Set} network is illustrated in Fig. \ref{fig:Architecture}. In the proposed architecture, all convolution (and deconvolution) operations are applied along the temporal dimension of the feature maps, automatically uncovering temporal signal patterns within the time span of the filters.



In order to provide a clear formulation of the problem, here we introduce the notations used throughout this paper. In this paper, we use $\mathcal{Y}$ for a set with unknown cardinality and $\mathcal{Y}^{m}$ for a set with known cardinality $m$. We define the set of $M$ supported activity elements by $\mathcal{A}=\{a_i\}_{i=1}^{M}$. Consider a collected data stream which contains raw time-series recordings from $d$ distinct sensor channels. We assume that for a subset of the recordings, sample annotation is not provided. Accordingly, adopting time-series segmentation with a sliding window size of $w$ on the data stream results in: 

\begin{itemize}
    \item A labeled training dataset $\mathcal{S}=\{(\bm{x}_i,\mathcal{Y}_i^{m_i})\}_{i=1}^{N_1}$ of size $N_1$, where each training instance is a pair consisting of a sensory segment $\bm{x}_i\in \mathbb{R}^{d\times w}$ with a fixed 2D representation and a target activity set $\mathcal{Y}_i^{m_i} = \{a_1,\ldots,a_{m_i}\} \subseteq \mathcal{A}, \left\vert{\mathcal{Y}_i}\right\vert = m_i, m_i \in \mathbb{Z}^{+} $.
    \item  An unlabeled dataset $\mathcal{V}=\{\bm{\bar{x}}_i\}_{i=1}^{N_2}$ of size $N_2$, where each instance is an unlabeled sensory segment $\bm{\bar{x}}_i\in \mathbb{R}^{d\times w}$.
\end{itemize}

 In order to leverage a larger number of segments for the unsupervised feature learning task, we define the unlabeled training dataset $\mathcal{U}=\{\bm{x^{\prime}}_i\}_{i=1}^{N_1+N_2}=\mathcal{V}\cup\{\bm{x}_i\}_{i=1}^{N_1}$ where each training instance $\bm{x^{\prime}}_i\in \mathbb{R}^{d\times w}$ is either a segment whose target activity set was not provided in the first place or a segment whose target set was intentionally discarded to augment the unlabeled dataset. 

\subsection{Unsupervised Feature Learning}\label{sub:unsup}
Through stacked hidden layers of encoding-decoding operations, auto-encoder learns latent representations of the sensory data in an unsupervised fashion. The reconstruction of unlabeled segments captures the process in which the sensor signals are generated and allows for the correlations between various sensor channels to be captured. Thus, the latent representations learned by the auto-encoder serve as efficient features that are highly effective in discriminating activity patterns. Formally, the input to the convolutional auto-encoder network is an unlabeled sensory time segment $\bm{x^{\prime}}\in\mathcal{U}$ on which the encoder network $f_{\text{enc}}: \mathbb{R}^{d\times w}\to\mathbb{R}^{p} $ (parameterized by $\Theta_{\text{enc}}$) is firstly applied to obtain the latent representation $\bm{z}_{\bm{x}^{\prime},\Theta_{\text{enc}}}$, \emph{i.e.} \begin{equation}
\bm{z}_{\bm{x}^{\prime},\Theta_{\text{enc}}} = f_{\text{enc}}(\bm{x^{\prime}};\Theta_{\text{enc}}).
\label{eq:z}
\end{equation}The resulting latent representation $\bm{z}_{\bm{x}^{\prime},\Theta_{\text{enc}}}\in\mathbb{R}^{p}$ is then utilized by the decoder network $f_{\text{dec}}: \mathbb{R}^{p}\to\mathbb{R}^{d\times w}$ (parameterized by $\Theta_{\text{dec}}$) to reconstruct the input. Noting that the generated reconstruction is directly influenced by the values of $\Theta_{\text{enc}}$ and $\Theta_{\text{dec}}$, we define the loss incurred by the output of auto-encoder network (illustrated by the dashed path in Fig. \ref{fig:Architecture}) given the unlabeled segment $\bm{x^{\prime}}$ as
\begin{equation}
\mathcal{L}_{\text{auto}}(\bm{x^{\prime}};\Theta_{\text{enc}},\Theta_{\text{dec}})= {\lVert \bm{x^{\prime}}- f_{\text{dec}}(\bm{z}_{\bm{x}^{\prime},\Theta_{\text{enc}}};\Theta_{\text{dec})} \rVert}^{2}.
\label{eq:L_auto}
\end{equation} We adopt an end-to-end approach towards training the convolutional auto-encoder parameters by minimizing the reconstruction objective on the unlabeled dataset $\mathcal{U}$
\begin{equation}
(\Theta^{*}_{\text{enc}},\Theta^{*}_{\text{dec}})=\arg\min_{\Theta_{\text{enc}},\Theta_{\text{dec}}}\sum_{i=1}^{N_1+N_2}\mathcal{L}_{\text{auto}}(\bm{x^{\prime}}_i;\Theta_{\text{enc}},\Theta_{\text{dec}}).
\label{eq:opt_solauto}
\end{equation}
In this architecture, the encoder network extracts features from unlabeled data and the decoder network uses the learned features to reconstruct the input. As the unsupervised training process progresses and the corresponding reconstruction loss is reduced, the network uncovers better feature representations of the sensory data. As a result, the acquired encoder network weights ($\Theta_{\text{enc}}^{*}$) can later be adopted in favor of a better guided supervised training.   

\subsection{Supervised Activity Set Learning and Inference} \label{sub:sup}

Using the labeled training dataset $\mathcal{S}=\{(\bm{x}_i,\mathcal{Y}_i^{m_i})\}_{i=1}^{N_1}$, the goal here is to train an activity set network that predicts a set of activity elements $\mathcal{Y^{+}}=\{a_1,\ldots,a_m\}$ with unknown and unfixed cardinality $m$ for a given test sensor segment $\bm{x}^+$. In our architecture, this is carried out by optimizing a \textit{set objective} through tuning the activity set network parameters which include weights corresponding to the encoder layers ($\Theta_{\text{enc}}$) as well as the extra dense layers ($\Omega$) in the classification head. Similar to \cite{Rezatofighi:2017}, in this paper we adopt joint learning and inference to learn and predict activity sets for HAR which we describe in what follows.



\subsubsection{Set Learning}\label{sec:learning}
In order to develop an accurate HAR system that meets the application demands, the network is required to correctly predict both the set cardinality (number of ongoing activities) as well as the set elements (activity types) given a sensory segment. Formally, given an input segment $\bm{x}$, the output of our activity set network comprises of: i) a \textit{set cardinality} term $f_{m^{\prime}}(\bm{x})$ with log softmax activation which produces cardinality scores; as well as ii) a \textit{set element} term $f_{a^{\prime}}(\bm{x})$ with sigmoid activation which produces scores for the set elements (activity types). In order to compute the loss incurred by the output of the activity set network (shown by the solid path in Fig. \ref{fig:Architecture}) given a labeled segment $\bm{x}$ with the target set $\mathcal{Y}^m$, we define our set objective as \begin{equation}
\begin{aligned}
\mathcal{L}_{\text{set}}(\bm{x},\mathcal{Y}^m;\Theta_{\text{enc}},\Omega) =&  
\sum_{a\in\mathcal{Y}}\ell_{\text{bce}}(a,f_{a^{\prime}}(\bm{x};\Theta_{\text{enc}},\Omega))\\
&+\ell_{\text{nll}}\left(m,f_{m^{\prime}}(\bm{x};\Theta_{\text{enc}},\Omega\right)),
\label{eq:L_set}
\end{aligned}
\end{equation} where $\ell_{\text{nll}}$ and $\ell_{\text{bce}}$ denote the negative log likelihood loss and the binary cross entropy loss, respectively. We consider the same \emph{i.i.d} assumption adopted in \cite{Rezatofighi:2017} for the set elements and perform MAP estimate to train the network parameters by minimizing the set objective on the labeled dataset $\mathcal{S}$, \emph{i.e.}
\begin{equation}
(\Theta^{*}_{\text{enc}},\Omega^{*})=\arg\min_{\Theta_{\text{enc}},\Omega}\sum_{i=1}^{N_1}\mathcal{L}_{\text{set}}(\bm{x}_i,\mathcal{Y}_i^{m_i};\Theta_{\text{enc}},\Omega).
\label{eq:opt_set}
\end{equation}
As such, $\Theta^{*}_{\text{enc}}$ and $\Omega^{*}$ are estimated by computing the partial derivatives of the objective function in Eq.~(\ref{eq:L_set}) and employing standard backpropagation in order to learn the network parameters.  

\subsubsection{Set Inference}
\label{sec:inference}
During the prediction phase for a given time segment $\bm{x}^+$, the goal is to predict the most likely set of activity elements $\mathcal{Y}^{*}=\{a_1,\ldots,a_m\}$. Using the optimal parameters ($\Theta^{*}_{\text{enc}},\Omega^{*}$) learned from the training dataset $\mathcal{S}$, a MAP inference is adopted to output the most likely activity set as
\begin{equation}
\begin{aligned}
\mathcal{Y}^{*} 
& = \arg\max_{m^{\prime},\mathcal{Y}^{m^{\prime}}}  \enspace f_{m^{\prime}}\left(\bm{x}^+;\Theta_{\text{enc}}^*,\Omega^*\right)+ m^{\prime}\log U\\
& + \sum_{a^{\prime}\in\mathcal{Y}^{m^{\prime}}} \log f_{a^{\prime}}(\bm{x}^{+};\Theta_{\text{enc}}^*,\Omega^*),
\end{aligned}
\label{eq:inference}
\end{equation}
where $U$, estimated from the validation set of the data, is a normalization constant that allows comparison between sets with different cardinalities. We derive the optimal solution for the above problem by solving a simple linear program as suggested in ~\cite{Rezatofighi:2017}.

\section{Experiments}


\subsection{Datasets}

For the evaluation of our approach, we adopt two widely used public HAR datasets that present both periodic and static activities. These benchmarks are elaborated as follows:

\begin{itemize}
	\item \textbf{WISDM Actitracker dataset} \cite{Kwapisz:2011}: This dataset contains 1,098,207 triaxial accelerometer readings gathered from 36 users which reflect activity patterns of \emph{walking}, \emph{jogging}, \emph{sitting}, \emph{standing}, and \emph{climbing stairs}. The acceleration measurements are collected with Android mobile phones at a constant sampling rate of 20 Hz. We randomly select recordings from 8 users as the testing set and use the remaining data as our training and validation sets. 
	
	
	\item \textbf{Opportunity dataset} \cite{chavarriaga:2013}: This dataset comprises annotated recordings from a wide variety of on-body sensors configured on four subjects while carrying out morning activities. The annotations include several modes of locomotion along with a \emph{Null} activity (referring to non-relevant activities) which makes the recognition problem much more challenging. For data collection, subjects were instructed to perform five Activities of Daily Living (ADL) runs as well as a drill session with 20 repetitions of a predefined sequence of activities. Each sample in the resulting dataset corresponds to 113 real valued signal measurements recorded with a sampling rate of 30 Hz. We employ the same subset of data as in the Opportunity challenge \cite{chavarriaga:2013} for training and testing purposes: ADL runs 4 and 5 collected from subjects 2 and 3 compose our testing set, and the remainder of the recordings from subjects 1,2 and 3 form our training and validation sets. 
\end{itemize}

\subsection{Data Preparation}

The preparation process involves performing per channel normalization to scale real valued attributes to [0,1] interval as well as segmentation and ground-truth generation, as described below.

\noindent\textbf{Time-series Segmentation:} Following the experiments in \cite{Kwapisz:2011,Alsheikh:2016}, we fix the sliding window size $w$ to incorporate 200 samples for both datasets (i.e, segments of 10 and 6.67 seconds duration for Actitracker and Opportunity dataset, respectively). However, since using non-overlapping sliding windows hinders real-time recognition of human activities, we set the sliding window stride to 20 samples. Such a deployment setting leads to generating predictions every second for the Actitracker dataset and every 0.67 seconds for the Opportunity dataset. 

\noindent\textbf{Set Ground-Truth Preparation:} Considering the sample annotations of a windowed sensory excerpt, the goal is to prepare the corresponding target set of activity elements as the training data. To this end, we consider a minimum \textit{expected recognition length} denoted by $r$, based on which we include activities in the target set. As such, if a minimum of $r$ sample annotations from a specific activity are observed in a time segment, the activity label appears in the target set. If no activity persists for the duration of $r$, the target activity set is considered as an empty set $\{\}$, representing the Null activity segment. In our experiments, we set $r$ to half the sensor sampling rates; \emph{i.e.}, 10 and 15 for Actitracker and Opportunity datasets, respectively.

\subsection{Evaluation Metrics} 

We employ the widely used HAR evaluation measures to report the performance of the baselines and our \emph{Deep Auto-Set} network. We select per-label \textit{precision} ($P$), \textit{recall} ($R$) and \textit{F1-score} ($F1$) instead of accuracy since accuracy is a bias estimator in the presence of class imbalance. For a specific activity label, label-based precision is defined as the ratio of the correctly predicted label occurrences over the total number of label occurrences in the predictions. Similarly, per-label recall is defined as the ratio of the correctly predicted label occurrences over the total number of label occurrences in the ground-truth. In this regard, per-label F1-score corresponds to the harmonic mean of precision and recall. Accordingly, $P_{\text{mean}}$, $R_{\text{mean}}$ and $F_{\text{mean}}$ are calculated by averaging across the per-label measures. 

We also use the overall \textit{exact match ratio} ($\textrm{MR}$), as adopted in \cite{Guo:2011,Alessandro:2013}, to report a harsh evaluation of performance. This metric requires the predicted activity set to exactly match the corresponding target set (both in terms of the set cardinality and the set elements) and therefore, does not tolerate partially correct predictions. For instance, no credit is considered for a predicted set of \{walk\} when the target set is \{walk,stand\}. We further decompose this measure over different activity set cardinalities $c$ and additionally report $\textrm{MR}_c$; \emph{i.e}, for instance $\textrm{MR}_2$ corresponds to the number of correctly predicted activity sets with cardinality of 2 over the total number of target sets with this cardinality.

\subsection{Implementation Details}

In this paper, the experiments are implemented using Pytorch \cite{Paszke:2017} as the deep learning framework and are run on a machine with a single GPU (NVIDIA GeForce GTX 1060). The network parameters are learned using ADAM optimizer with weight decay and initial learning rate respectively set to $5\cdot 10^{-5}$ and $10^{-4}$, on mini-batches of size 64 by backpropagating the gradients of corresponding loss functions. For the supervised training step, the optimizer learning rate is scheduled to gradually decrease after each epoch. Moreover, training is stopped if validation objective does not decrease for 5 subsequent epochs. Accordingly, the corresponding weights for the epoch with the best validation performance are applied to report performance on the testing sets. The hyper-parameter $U$ is set to be 2.5 and 3.4, respectively adjusted on the validation sets of Actitracker and Opportunity datasets. We refer interested readers to \cite{Hammerla:2016} for excellent guidelines on setting architecture and optimizer hyper-parameters. 

\subsection{Results}

\begin{table*}[]
	\centering
	\caption{Performance evaluation of the baseline CNN architecture \protect\cite{Ordonez:2016} trained with multi-class formulated objective against both the approximated ground truth (equivalent to the last observed sample annotation) as well as the actual ground-truth for Opportunity dataset.}
	\label{tab:1}
	\begin{tabular}{c|c|c|c|c}
		\hline
		Model  & Network Prediction & Evaluation Ground Truth   & $F_{\text{mean}}$ & $\textrm{MR}$  \\ \hline
		\multirow{2}{*}{CNN \cite{Ordonez:2016}}& \multirow{2}{*}{Single activity label}
		& Last sample's label  & 0.890  &  87.4$\%$   \\ 
		&      & Actual labels & 0.793  &  54.7$\%$   \\ 
		\hline
	\end{tabular}
\end{table*}

\begin{table*}[t]
	\centering
	\caption{Comparison of our proposed \emph{Deep Auto-Set} network against the baselines according to the obtained exact match ratio for each dataset. The best results are highlighted with boldface. Note that for the Actitracker dataset, sensor segments with cardinality of 0 (corresponding to Null segments) and 3 do not exist.}
	\label{tab:3}
	\begin{tabular}{c|l|c|cccc}
		\hline
		Dataset  & Model   & $\textrm{MR}$ & $\textrm{MR}_{0}$ & $\textrm{MR}_{1}$ & $\textrm{MR}_{2}$ & $\textrm{MR}_{3}$  \\ \hline
		\multirow{4}{*}{Actitracker} 
		& (Baseline) Deep-BCE  & 90.1$\%$  &   -    &   91.1$\%$   & 60.2$\%$   & -     \\ 
		& (Ours) Auto-BCE & 92.9$\%$   &   -    &   93.9$\%$   & 62.7$\%$   & - \\ 
		& (Ours) Deep-Set & 93.2$\%$   &   -    &   93.9$\%$   & 71.5$\%$   & - \\ 
		& \textbf{(Ours) Auto-Set} & \textbf{94.9}$\textbf{\%}$  &  -   &   \textbf{95.5}$\textbf{\%}$ & \textbf{75.1}$\textbf{\%}$ & - \\ \hline
		\multirow{4}{*}{\begin{tabular}[c]{@{}c@{}}Opportunity\\ (locomotions)\end{tabular}} 
		& (Baseline) Deep-BCE  & 82.0$\%$  &  70.7$\%$   & 85.0$\%$  &   84.9$\%$      & 68.3$\%$     \\ 
		& (Ours) Auto-BCE  & 83.1$\%$  &  73.7$\%$   & 85.1$\%$  &   85.3$\%$      & 69.9$\%$ \\ 
		& (Ours) Deep-Set  & 83.9$\%$  &  78.2$\%$   & 86.8$\%$  &   84.9$\%$      & 68.7$\%$ \\ 
		& \textbf{(Ours) Auto-Set}  & \textbf{84.9}$\textbf{\%}$ &   \textbf{80.2}$\textbf{\%}$  &    \textbf{87.1}$\textbf{\%}$ & \textbf{85.6}$\textbf{\%}$ & \textbf{75.6}$\textbf{\%}$ \\ \hline
	\end{tabular}
\end{table*}

\begin{table}[t]
	\centering
	\caption{Comparison of our proposed \emph{Deep Auto-Set} network against the baselines according to the obtained mean F1-score ($F_{\text{mean}}$), precision ($P_{\text{mean}}$) and recall ($R_{\text{mean}}$) for each dataset. The best results are highlighted with boldface.}
	\label{tab:2}
	\begin{tabular}{c|l|c|c|c}
		\hline
		Dataset  & Model   & $F_{\text{mean}}$ & $P_{\text{mean}}$ & $R_{\text{mean}}$  \\ \hline
		\multirow{4}{*}{Actitracker} 
		& (Baseline) Deep-BCE  & 0.943  &  0.908  &   0.980     \\ 
		& (Ours) Auto-BCE  & 0.966  &  0.949  &   0.983 \\ 
		& (Ours) Deep-Set  & 0.961  &  0.943  &   0.980 \\ 
		& \textbf{(Ours) Auto-Set}  & \textbf{0.973}   & \textbf{0.957} & \textbf{0.989} \\ \hline
		\multirow{4}{*}{\begin{tabular}[c]{@{}c@{}}Opportunity\\ (locomotions)\end{tabular}}  
		& (Baseline) Deep-BCE  & 0.927  &  0.901  &   0.954     \\ 
		& (Ours) Auto-BCE  & 0.936  &  0.918  &   0.955 \\ 
		& (Ours) Deep-Set  & 0.934  &  0.915  &   0.955 \\ 
		& \textbf{(Ours) Auto-Set}  & \textbf{0.943}   & \textbf{0.927} & \textbf{0.960} \\ \hline
	\end{tabular}
\end{table}

\begin{figure*}[h]
	\centering
	\vspace{8mm}
	\includegraphics[width=120mm, keepaspectratio]{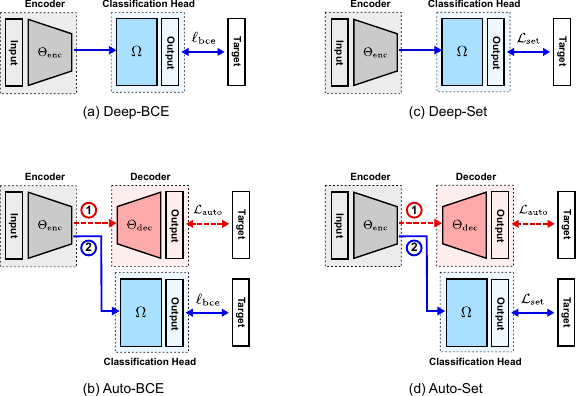}
	\caption{An overview of different activity recognition models explored in this paper. Note that all models adopt the same network architecture to generate classification outputs and thus, share the same number of parameters. Therefore, the enhanced recognition performance is a product of effective unsupervised feature learning as well as incorporating novel set loss function for the underlying problem.}
	\label{fig:Baselines}
\end{figure*}

A key motivation for our work is the activity information loss that is incurred by conventional ground truth approximations in multi-class problem formulations. In order to verify this, we conform to the conventional multi-class formulation of HAR and train the CNN in~\cite{Ordonez:2016} by minimizing the multi-class classification objective. In Table \ref{tab:1}, we report performance of the resulting HAR system by comparing the generated predictions against both the \textit{approximate} ground truth (obtained from the last observed sample annotation) as well as the \textit{actual} multi-label ground truth for Opportunity dataset. To clarify, consider the scenario where a sensory segment of interest initiates with measurements of walking and terminates with standing. Thus, the approximate ground truth would be \emph{standing} but the actual ground truth labels are the set \{\emph{walking},\emph{standing}\}. Assuming that the network solving the multi-class formulated problem predicts the underlying activity to be \emph{standing}, in our evaluation against the actual ground truth represented by the set of labels \{\emph{walking},\emph{standing}\}, the predicted class \emph{standing} is treated as a true positive whereas the missing class \emph{walking} is considered as a false negative. 

In Table \ref{tab:1}, the lower performance measures obtained from the evaluation against the actual ground truth labels as compared with the approximated ground truth suggest that there are sensory segments in the HAR dataset that convey measurements of multiple activities in the time span of the sliding window---see the result for MR in Table~\ref{tab:1}. For these segments, approximating the ground-truth can lead to missed activity information for a  multi-class formulation of HAR, especially in the presence of short duration activities such as activity transitions~\cite{yao2018efficient}. In contrast, a set-based formulation allows capturing the presence of multiple activity labels in the ground truth. 
Although we have shown a comparison for a multi-class problem formulation commonly employed for HAR, we can see that it is not possible to make a fair comparison with our set-based formulation beyond what we have observed here. Therefore, we omit empirical comparisons with existing multi-class based solutions and instead present evaluation against multi-label based activity recognition systems that can handle segments with multiple activities.
\vspace{3mm}

\noindent\textbf{Activity Recognition Models:} Fig. \ref{fig:Baselines} illustrates the schematic architectures for:
\begin{itemize}
    \item \textit{Deep-BCE}: A conventional multi-label model that follows a purely supervised minimization of binary cross entropy loss ($\ell_{\text{bce}}$) for training and heuristic thresholding of activity scores for inference.
    \item \textit{Auto-BCE}: A conventional multi-label model that leverages a prior unsupervised feature learning step via minimization of reconstruction objective ($\mathcal{L}_{\text{auto}}$) as well as a supervised optimization of binary cross entropy loss.
    \item \textit{Deep-Set}: A set-based model that follows a purely supervised optimization of the set objective ($\mathcal{L}_{\text{set}}$) proposed in Eq. \eqref{eq:L_set} for training and the MAP inference introduced in Eq. \eqref{eq:inference} for set inference.
    \item \textit{Auto-Set}: The proposed Deep Auto-Set model elaborated in Section \ref{method}.
\end{itemize}   

Notably, as opposed to existing multi-class based HAR systems which are restricted to predict a single activity class even when an activity transition takes place within a segment, all recognition models adopted in this paper are capable of predicting multiple activities for a given sensory segment. We adopt the same layer operations presented in Fig. \ref{fig:Architecture} for supervised and unsupervised training steps of the baseline models.



The performance results of our \emph{Deep Auto-Set} network and the baseline models on the two HAR representative datasets are shown in Table \ref{tab:3} and Table \ref{tab:2} for different evaluation metrics. From the reported results, we can see that our novel \emph{Deep Auto-Set} network consistently outperforms the baselines on Actitracker and Opportunity datasets in terms of both F1-score and exact match ratio performance metrics. 
Moreover, the match ratios in Table \ref{tab:3} suggest that \emph{Deep Auto-Set} is a robust activity recognition system capable of: $i$) distinguishing different activity classes accurately (implied from $\textrm{MR}_0$ and $\textrm{MR}_1$ values); $ii$) identifying activity transition segments (implied from $\textrm{MR}_2$ values); as well as $iii$) recognizing short duration human activities (implied from $\textrm{MR}_3$ values).

We summarize the experimental results on both datasets by concluding that:
\begin{itemize}
	\item Activity recognition systems that leverage unlabeled data present better performance over their solely supervised variants; \emph{e.g.}, note the improved performance of \emph{Auto-BCE} over \emph{Deep-BCE}.
	\item Compared with a conventional multi-label formulation: i) incorporating set loss into the training process can allow the network to learn mulitple activities represented in the ground truth data of a given segment more accurately; and ii) the set inference procedure can jointly exploit cardinality and set element scores to generate predictions instead of empirically determined thresholds; \emph{e.g.}, note the performance improvement of \emph{Deep-Set} over \emph{Deep-BCE}. 
	\item While each component of our proposed methodology (unsupervised feature learning and supervised set learning) individually introduces performance boost in recognition of human activities, when coupled together in a unified framework, the resulting HAR system proves to be the most effective.
\end{itemize}




\section{Conclusions}
In this paper, we defined human activity recognition as a set prediction problem. In contrast to the conventional multi-class treatment of HAR problems, our intuitive formulation allows sensory segments to be associated with a set of activities and thus, naturally handles segments with multiple activities. In a unified architecture, we addressed the HAR problem by developing a deep HAR system that: i) exploits unlabeled data to uncover effective feature representations; and ii) incorporates a set objective to learn mappings from input sensory segments to target activity sets. To provide insights on how each component of our proposed methodology contributes to enhance recognition performance in isolation, we explored three different multi-label activity recognition models as our baselines. Finally, through empirical experiments on HAR representative datasets, we demonstrated the effectiveness of our proposed \emph{Deep Auto-Set} network for human activity recognition. 

While not explored in this paper, our proposed set-based methodology potentially offers an elegant solution for the challenging problem of concurrent human activity recognition. In this problem, the goal is to recognize not only the sequential but also the co-occurring activities from raw sensory time-series data. As a future direction to our current study, we intend to further investigate recognition of concurrent human activities.



\bibliographystyle{elsarticle-num}
\bibliography{main}

\begin{thebibliography}{10}
\expandafter\ifx\csname url\endcsname\relax
  \def\url#1{\texttt{#1}}\fi
\expandafter\ifx\csname urlprefix\endcsname\relax\def\urlprefix{URL }\fi
\expandafter\ifx\csname href\endcsname\relax
  \def\href#1#2{#2} \def\path#1{#1}\fi

\bibitem{Bao:2004}
L.~Bao, S.~S. Intille, Activity recognition from user-annotated acceleration
  data, in: A.~Ferscha, F.~Mattern (Eds.), Proceedings of the 2nd International
  Conference on Pervasive Computing, 2004, pp. 1--17.

\bibitem{Ordonez:2016}
F.~J. Ordóñez, D.~Roggen, Deep convolutional and lstm recurrent neural
  networks for multimodal wearable activity recognition, Sensors 16~(1).

\bibitem{Hammerla:2016}
N.~Y. Hammerla, S.~Halloran, T.~Pl\"{o}tz, Deep, convolutional, and recurrent
  models for human activity recognition using wearables, in: Proceedings of the
  25th International Joint Conference on Artificial Intelligence, 2016, pp.
  1533--1540.

\bibitem{yao2018efficient}
R.~Yao, G.~Lin, Q.~Shi, D.~C. Ranasinghe, Efficient dense labelling of human
  activity sequences from wearables using fully convolutional networks, Pattern
  Recognition 78 (2018) 252--266.

\bibitem{Bulling:2014}
A.~Bulling, U.~Blanke, B.~Schiele, A tutorial on human activity recognition
  using body-worn inertial sensors, ACM Computing Surveys 46~(3) (2014)
  33:1--33:33.

\bibitem{Kim:2010}
E.~Kim, S.~Helal, D.~Cook, Human activity recognition and pattern discovery,
  IEEE Pervasive Computing 9~(1) (2010) 48--53.

\bibitem{Yang:2015}
J.~B. Yang, M.~N. Nguyen, P.~P. San, X.~L. Li, S.~Krishnaswamy, Deep
  convolutional neural networks on multichannel time series for human activity
  recognition, in: Proceedings of the 24th International Conference on
  Artificial Intelligence, 2015, pp. 3995--4001.

\bibitem{Rezatofighi:2017}
S.~H. Rezatofighi, A.~Milan, Q.~Shi, A.~R. Dick, I.~D. Reid, Joint learning of
  set cardinality and state distribution, in: AAAI, 2018, (to appear).

\bibitem{Ravi:2005}
N.~Ravi, N.~Dandekar, P.~Mysore, M.~L. Littman, Activity recognition from
  accelerometer data, in: Proceedings of the 17th Conference on Innovative
  Applications of Artificial Intelligence, 2005, pp. 1541--1546.

\bibitem{Huynh:2005}
T.~Huynh, B.~Schiele, Analyzing features for activity recognition, in:
  Proceedings Conference on Smart Objects and Ambient Intelligence: Innovative
  Context-aware Services: Usages and Technologies, 2005, pp. 159--163.

\bibitem{Zhang:2012}
M.~Zhang, A.~A. Sawchuk, Motion primitive-based human activity recognition
  using a bag-of-features approach, in: Proceedings of the 2nd ACM SIGHIT
  International Health Informatics Symposium, 2012, pp. 631--640.

\bibitem{wickramasinghe2017sequence}
A.~Wickramasinghe, D.~C. Ranasinghe, C.~Fumeaux, K.~D. Hill, R.~Visvanathan,
  Sequence learning with passive rfid sensors for real-time bed-egress
  recognition in older people, IEEE Journal of Biomedical and Health
  Informatics 21~(4) (2017) 917--929.

\bibitem{Bulling:2012}
A.~Bulling, J.~A. Ward, H.~Gellersen, Multimodal recognition of reading
  activity in transit using body-worn sensors, ACM Transactions on Applied
  Perception 9~(1) (2012) 2:1--2:21.

\bibitem{Lara:2012}
O.~D. Lara, A.~J. P{\'e}rez, M.~A. Labrador, J.~D. Posada, Centinela: A human
  activity recognition system based on acceleration and vital sign data,
  Pervasive and Mobile Computing 8~(5) (2012) 717--729.

\bibitem{ShinmotoTorres:2017CRF}
R.~L. Shinmoto~Torres, Q.~Shi, A.~van~den Hengel, D.~C. Ranasinghe, A
  hierarchical model for recognizing alarming states in a batteryless sensor
  alarm intervention for preventing falls in older people, Pervasive Mob.
  Comput. 40~(C) (2017) 1--16.

\bibitem{Randell:2000}
C.~Randell, H.~Muller, Context awareness by analysing accelerometer data, in:
  Proceedings of the 4th International Symposium on Wearable Computers, 2000,
  pp. 175--176.

\bibitem{Lecun:2015}
Y.~LeCun, Y.~Bengio, G.~Hinton, Deep learning, Nature 521~(7553) (2015)
  436--444.

\bibitem{Zeng:2014}
M.~Zeng, L.~T. Nguyen, B.~Yu, O.~J. Mengshoel, J.~Zhu, P.~Wu, J.~Zhang,
  Convolutional neural networks for human activity recognition using mobile
  sensors, in: Proceedings of the 6th International Conference on Mobile
  Computing, Applications and Services, 2014, pp. 197--205.

\bibitem{Alsheikh:2016}
M.~A. Alsheikh, A.~Selim, D.~Niyato, L.~Doyle, S.~Lin, H.~Tan, Deep activity
  recognition models with triaxial accelerometers, in: Artificial Intelligence
  Applied to Assistive Technologies and Smart Environments, 2016.

\bibitem{Erhan:2010}
D.~Erhan, Y.~Bengio, A.~Courville, P.-A. Manzagol, P.~Vincent, S.~Bengio, Why
  does unsupervised pre-training help deep learning?, Journal of Machine
  Learning Research 11 (2010) 625--660.

\bibitem{Kwapisz:2011}
J.~R. Kwapisz, G.~M. Weiss, S.~A. Moore, Activity recognition using cell phone
  accelerometers, ACM SigKDD Explorations Newsletter 12~(2) (2011) 74--82.

\bibitem{chavarriaga:2013}
R.~Chavarriaga, H.~Sagha, A.~Calatroni, S.~T. Digumarti, G.~Tröster, J.~del
  R.~Millán, D.~Roggen, The opportunity challenge: A benchmark database for
  on-body sensor-based activity recognition, Pattern Recognition Letters
  34~(15) (2013) 2033 -- 2042.

\bibitem{Guo:2011}
Y.~Guo, S.~Gu, Multi-label classification using conditional dependency
  networks, in: Proceedings of the 22nd International Joint Conference on
  Artificial Intelligence, 2011, pp. 1300--1305.

\bibitem{Alessandro:2013}
A.~Alessandro, G.~Corani, D.~Mau{\'a}, S.~Gabaglio, An ensemble of bayesian
  networks for multilabel classification, in: Proceedings of the 23rd
  International Joint Conference on Artificial Intelligence, 2013, pp.
  1220--1225.

\bibitem{Paszke:2017}
A.~Paszke, S.~Gross, S.~Chintala, G.~Chanan, E.~Yang, Z.~DeVito, Z.~Lin,
  A.~Desmaison, L.~Antiga, A.~Lerer, Automatic differentiation in pytorch.

\end{thebibliography}

\end{document}